# Accountability in Autonomous Drone-Based Firefighting: Insights From a Field Trial

*Completed Research Paper*

**Dzmitry Katsiuba**
University of Zurich
Zurich, Switzerland
katsiuba@ifi.uzh.ch

**Anna Katharina Boos**
University of Zurich
Zurich, Switzerland
anna.boos@uzh.ch

**Robin Hany**
University of Zurich
Zurich, Switzerland
robin.hany@uzh.ch

**Mateusz Dolata**
Zeppelin University
Friedrichshafen, Germany
mateusz.dolata@zu.de

**Gerhard Schwabe**
University of Zurich
Zurich, Switzerland
schwabe@ifi.uzh.ch

## Abstract

*There is a growing research field exploring how autonomous drones can enhance emergency response effectiveness. Integrating these (artificial) agents into existing emergency teams and workflows may significantly impact established accountability relationships. This paper examines how autonomous drones affect accountability attribution within complex socio-technical systems. Drawing on two real-life field trials in firefighting, the study reveals substantial uncertainty around accountability when drones are organizationally deployed. Using Bovens' accountability framework, two challenges are identified: (1) uncertainty about the role of drones within hierarchical structures, leading to confused accountability ascriptions; and (2) new forms of human-drone interactions introducing additional accountability-relevant issues. Based on these insights, the paper proposes actionable recommendations to support the responsible integration of autonomous drones into firefighting operations without undermining accountability. These findings offer practical guidance for policymakers and contribute to further research on accountability in autonomous systems.*

## Introduction

Unmanned aerial vehicles (drones) are becoming increasingly common in various contexts, both public and private (e.g., military, law enforcement, disaster response, agriculture, and logistics). The integration of autonomous drones into firefighting operations is likely to reshape existing accountability relations, raising new questions about responsibility allocation in high-stakes scenarios. While much of the existing literature on drone accountability is primarily of a normative-theoretical nature (Di Nucci & Sio, 2016) , there remains a lack of practical guidance for emergency services. To address this gap, we adopt the Bovens' accountability framework (Bovens et al., 2014) to examine real-world accountability challenges arising from the use of autonomous drones in firefighting. Our study draws on two field trials with 95 participants along with in-depth interviews. Our findings reveal a variety of accountability challenges, predominantly concerning the role of drones within hierarchical structures and decision-making processes of firefighting organizations.

Based on these findings, we propose eight actionable guidelines to support the effective integration of autonomous drones into firefighting operations while strengthening accountability.

Understanding these accountability challenges requires situating them within the broader context of drones' expanding technical capabilities and their growing role in emergency services. As drones become increasingly sophisticated, their potential application is widening, from autonomous operations to synergistic partnerships with human actors in various scenarios. Unsurprisingly, emergency services, including firefighting, have begun to explore these capabilities. For example, obtaining a bird's-eye view of the fire site has traditionally been an essential but challenging task for firefighting teams. Recent work by Chang et al. (2022) shows that drones can be used to support commanders on the ground with real-time reports by equipping drones with cameras and object detection capabilities. This enhancement enables drones to expedite the initial phase and reduce pressure on ground personnel. Similarly, studies by Hein et al. (2017) and Merino et al. (2012) confirm the operational benefits of camera-equipped drones in firefighting scenarios. Other research focuses on different aspects of firefighting, such as preventing and detecting fires (Ambrosia et al., 2003; Roldán-Gómez et al., 2021; Yuan et al., 2015). Roldán-Gómez et al. (2021) show that drones can even be deployed as an early warning system for wildfires. As this body of work suggests, drones can be applied in various emergency scenarios, and their capabilities are deemed to grow.

With the integration of Artificial Intelligence (AI), drone capabilities can be significantly enhanced, enabling them to perform complex tasks either autonomously or in close collaboration with human teams. However, the deployment of such advanced technologies entails a range of ethical challenges. Compliance with moral, legal, and professional standards is essential to ensure long-term public legitimacy of drone use. To secure such compliance, robust accountability frameworks are crucial. Following Boven's theoretical framework (2007; 2014), accountability refers to an actor's obligation to explain and justify their conduct to a relevant forum, which, in turn, holds the authority to demand answers and impose consequences. Stakeholders, especially those directly affected, should be able to scrutinize AI-enabled operations to ensure that violations of moral, legal, and professional standards can be revealed and properly addressed (Kroll, 2020; Novelli et al., 2024). Accountability – a cornerstone of democratic and ethical governance – is vital in many contexts, however, it cannot be overstated in context of high-stakes situations such as emergency response where people's lives are at risk. Scholars are worried that the increasing use of AI may undermine established accountability relations (Busuioc, 2021; Lepri et al., 2018; Nissenbaum, 1996; Santoni de Sio & Mecacci, 2021). The introduction of AI in socio-technical organizations can shift the distribution of responsibility among team members, creating uncertainty about who is accountable when failures or accidents occur. Some are concerned that this situation invites deployers to use AI systems as moral scapegoats, thereby denying their own accountability (Nissenbaum, 1996).

Imagine the following scenario: *A fire broke out in a forest and firefighters are on their way. On site, the firefighter team deploys an autonomous drone to scan the area for victims and assess their current health status according to body temperature and movement. Based on the findings, the drone sends a list of all identified - possibly wounded - people to the commander in charge. Based on that information, the commander instructs the team in which order they should rescue the victims. With two critically wounded victims, the team makes its way to rescue the first person on the list. Now, it turns out that the first victim is a wounded animal that was mistaken for a human by the drone. With this delay, the second most critical victim does not receive the needed help in time.*

Although this is a fictitious scenario, one can easily imagine something similar happening if the application of autonomous drones in firefighting emergencies becomes more widespread. The scenario brings the following challenge to the surface: Who can and is to be held accountable for the drone's failure and the victim's loss? In other words, who must answer for which actions and decisions, and who ought to bear the consequences? Motivated by this challenge, we use Bovens' accountability framework (2007; 2014) to examine people's perceptions of autonomous drones and the associated accountability issues in emerging socio-technical systems. Our study is guided by the following research questions:

> *RQ1: How does the application of autonomous drones affect the attribution of accountability in a firefighting use case?*
>
> *RQ2: What measures can mitigate the associated problems in accountability attribution?*

Our aim is to investigate how accountability is attributed in real-life firefighting scenarios and to develop empirically informed recommendations to address emerging accountability challenges. This research is part of a project focused on exploring the potential applications of autonomous drones in firefighting, undertaken in close collaboration with University of Zurich and two Swiss fire departments. The project follows three key steps: (I) workshops and interviews with firefighters to generate ideas for implementing autonomous drone systems, (II) two field trials with local volunteer fire departments, and (III) interviews with various participants involved in the trials. The research team – comprising philosophers and computer scientists – brought diverse perspectives to the qualitative analysis and interpretation of the data. The findings provide a framework for understanding emerging accountability challenges and form the basis for the actionable guidelines that we present towards the end of this article.

The use of information systems in emergency services is highly relevant to information systems (IS) discipline. Emergency services and disaster response are a common topic in IS literature (Beydoun et al., 2018; Prasanna et al., 2017), indicating that such situations involve a complex set of interrelated activities that are both knowledge-intensive and urgent, forming a challenging setting for sociotechnical designs (Beydoun et al., 2018; Prasanna et al., 2017). The research has focused on the design (Prasanna et al., 2017; Turoff et al., 2004) and professional adoption (Weidinger et al., 2018, 2022) of emergency response information systems (ERIS), indicating, among other things, that systems supporting on-site dynamic information collection offer major advantages to first responders (Yang et al., 2009). Drones exemplify such on-site dynamic information systems. However, achieving this requires resolving multiple problems related to the public's acceptance of autonomous systems (Dolata & Aleya, 2022; Dolata & Schwabe, 2023). Given the rapid uptake of drones in practice and intense discourse in adjacent disciplines like medical informatics (Balasingam, 2017; Krey & Seiler, 2019) or human-computer interaction (Boonyard et al., 2025; Dolata & Aleya, 2022; Wu et al., 2025), we assume that autonomous drones will become important element of ERIS and crucial for emergency processes.

An important aspect is addressing issues around accountability and responsibility. Whereas IS has studied AI accountability extensively in the past (Dennehy et al., 2023; Nguyen et al., 2024), the research remains conceptual and abstracts from specific applications of AI and autonomous systems (Nguyen et al., 2024, 2024). This study contributes to this important IS discourse by delivering an empirically-based view on accountability perceptions using emergency services as the practical context and extending the existing theoretical perspective based on the concept of accountability in democratic theory.

# Related Work

## *The Concept of Accountability in Democratic Theory*

In political theory, accountability is widely regarded as a core principle of democratic legitimacy (Bovens et al., 2014; Mulgan, 2003; Warren, 2014). Public institutions, including emergency services, are vested with far-reaching competencies and authoritative powers over citizens, with potentially significant consequences for their well-being. This makes citizens vulnerable to flawed decision making, misconduct, and abuses of power. To mitigate these risks, accountability is key. As a pivotal democratic safeguard, accountability provides the public with the means to scrutinize, appeal, and – when necessary – sanction improper behavior, ensuring that public institutions work in the public's best interest. By holding public institutions accountable, the public can monitor and control government conduct and enhance the learning capacity and efficiency of public administrations (Bovens, 2007). Therefore, the ability and openness of public institutions to render an account to the public is an essential feature of the democratic exercise of power.

Bovens (2007, p. 450) defines accountability as "a relationship between an actor and a forum, in which the actor has an obligation to explain and to justify his or her conduct, the forum can pose questions and pass judgment, and the actor may face consequences" (see Figure 1). Bovens (2010) stresses that accountability should not be understood as a substantive ethical norm, but rather as an institutional mechanism for ensuring that actors take responsibility for their actions when established norms are violated. This is particularly important when operations do not go as expected and result in losses. Applied to our use case, "the actor" amounts to the fire brigade, a publicly funded institution with far reaching competences and thus subject to public scrutiny. Ensuring accountability – and, by extension, democratic legitimacy – requires the fire brigade to disclose operational details and justify their decisions when requested by the public, particularly by those directly affected. The forum usually refers to those with a legitimate claim to oversee the actor's decisions, ideally overlapping with those directly affected by these decisions. As to our case, this would specifically amount to the bystanders but includes the general public as well. Both, bystanders and the public in general, have a right to demand explanation and justification from the fire brigade, and call for consequences, such as sanctions or compensation. Oftentimes, the actors' facing consequences also includes the duty to take effective measures to prevent events from reoccurring (Thompson, 2017).

IS literature on accountability, though limited, often adopts Bovens' (2007) perspective, such as when examining intentions to violate access policies and developing interfaces that enhance users' perceptions of accountability if they do so (Vance et al., 2013, 2015). Similar take at accountability is put forward in studies around healthcare information systems (Sjöström et al., 2022). However, to explore information sharing behavior in collaboration with AI, a definition from O'Kelly and Dubnick (2019), which is specifically tailored for misuse and violation situations, was used. Overall, until recently, the accountability lens has primarily been used in IS studies to understand noncompliant behavior of IT users in organizations.

Simultaneously, an increasing number of recent IS papers explicitly address the topic of AI accountability (Kempton et al., 2023; Nguyen et al., 2024). While existing studies provide interesting insights into achieving AI accountability through specific system, process, and data design (Kempton et al., 2023; Kim et al., 2020), research on attributing accountability in scenarios where humans and AI systems form a single sociotechnical system, such as when firefighters and drones collaborate during an emergency, remains scarce. Given that high-stakes scenarios are more likely to involve human-AI teams rather than AI acting independently, it is imperative to explore accountability in these more complex situations empirically.

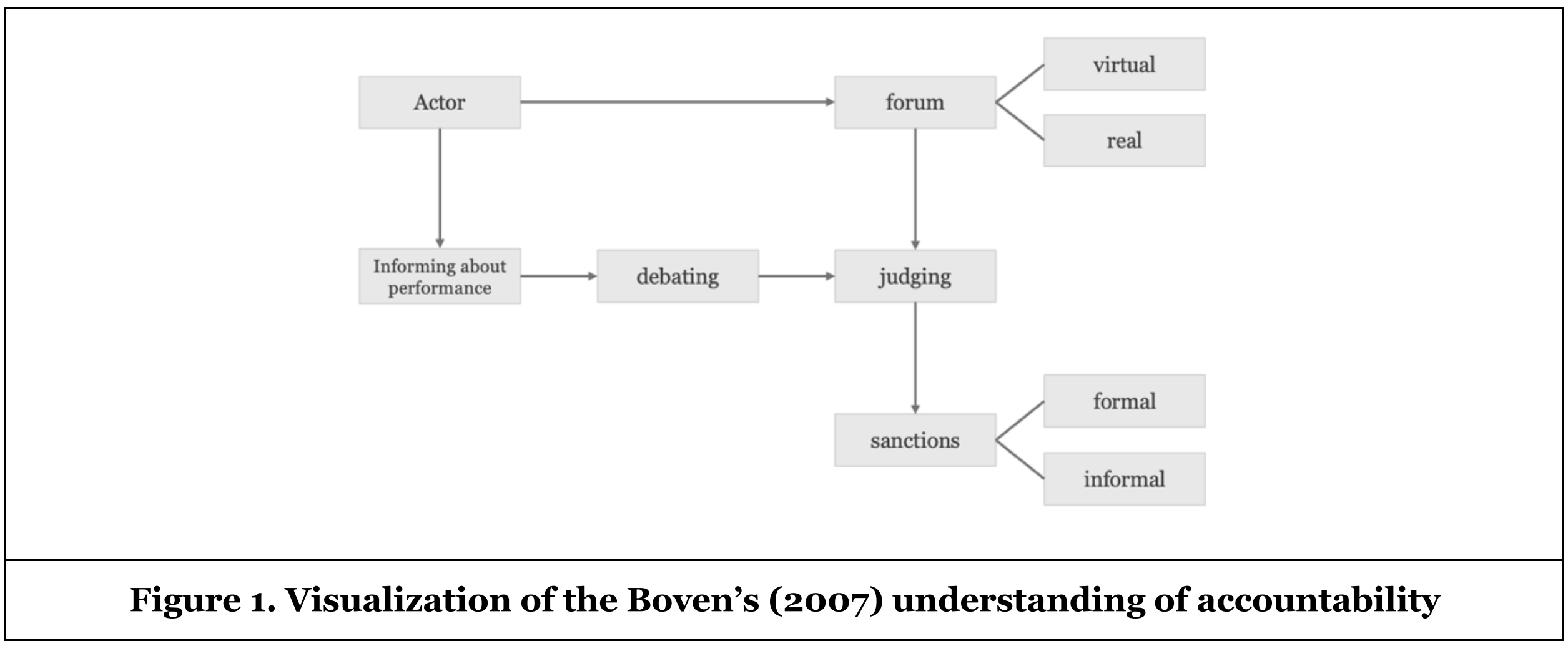

**Figure 1. Visualization of the Boven's (2007) understanding of accountability**

## *The accountability problem in AI*

Besides democratic theory, accountability is considered a key concept also in the governance and responsible use of AI. Kroll (2020) argues that accountability is a crucial governance concept for technical systems like AI because it specifies who is answerable to whom and for what, thereby linking complex and often opaque algorithmic processes to social, political, and moral norms. Unlike other abstract ethical values – such as fairness, justice, or privacy – which often resist explicit formalization and are difficult to hard-code into the systems, accountability offers a tractable way to structure oversight and enable responsible AI deployment. While Lechtermann (2022) agrees on the fundamental importance of accountability in AI governance, he cautions that it is not sufficient. Accountability should be understood as a tool to enforce agreed-upon norms, but its effectiveness depends on those norms being established and clearly articulated. Accountability cannot substitute for the political and ethical debate about what the substance of those norms ought to be.

In this respect, Kroll and Lechtermann both echo Bovens' understanding of accountability as an institutional mechanism to ensure compliance with existing moral, legal, and professional standards (Bovens, 2010). Yet it is precisely this mechanism that comes under pressure in the context of AI. The complexity and often opaque nature of AI can render it increasingly difficult to answer the two key questions with regard to accountability (Kroll, 2020; Lechterman, 2022): First, who is obligated and able to provide answers concerning the use and functionality of AI systems, and second, who must bear the consequences for undesired outcomes? There is the concern of "eroding accountability" in increasingly complex socio-technical systems, since it becomes ever more difficult to trace responsibility back to single parties and hold them accountable (Nissenbaum, 1996). As a result, actors may have an incentive to evade accountability by hiding behind the increasing socio-technical complexity and using technology as a scapegoat (Danaher, 2016).

Moreover, in the machine ethics literature, there is a vivid debate on so-called "accountability gaps." The issue was famously introduced by Matthias (2004) and has since been restated in various forms by de Jong (2020), Grote et al. (2020), and Sparrow (2007). As AI systems, such as autonomous drones, increasingly interact with their environment as independent agents and perform complex tasks once reserved for humans, the question arises: who ought to take responsibility when the drones' actions result in harm? Unlike humans, AI systems lack the fundamental metaphysical properties required to assume responsibility for their own conduct, even when they cause harm. At the same time, tracing responsibility back to developers, deployers, and operators proves equally problematic, since the very self-learning and autonomous features that define advanced AI systems inherently diminish human control over those systems. Hence, there may be no suitable actor to be held responsible and render an account: While AI systems are metaphysically incapable of doing so, human stakeholders cannot be legitimately held responsible for outcomes that lie beyond their control (Matthias, 2004). In response to this challenge, Santoni de Sio and van den Hoven (2018) argue that autonomous AI systems should be designed in ways that preserve the possibility of "meaningful human control", such that accountability-gaps can be avoided.

Finally, worries about the lack of control over AI are closely related to the lack of explainability and scrutability that is associated with algorithmic models, especially in the context of machine learning (Burrell, 2016; Busuioc, 2021; Cobbe et al., 2021). As a result, AI-systems are often "black-boxes" that may not "provide an explanation for their reasoning behind their outputs" (Binns, 2018, p. 553). Such "Black boxes" undermine the very bedrock of accountability – the ability to provide clear explanations and justifications for specific decisions and actions – thereby making it significantly harder to trace back accountability to the responsible individuals.

Despite the surge of interest in accountability within AI ethics, the topic remains notably underexplored in more technical, applied research. Much of the existing literature relies on argumentation or surveys (Aydin, 2019; Del-Real & Díaz-Fernández, 2021; Lin Tan et al., 2021; Miron et al., 2023), providing limited empirical evidence from real-world scenarios. This gap is particularly evident in studies of drones, which often focus on public acceptance in hypothetical situations rather than their actual use in practice. While some research examines the risks (Aydin, 2019) associated with drone applications – at various levels of impact and probability – these studies typically do not draw a connection to the concept of accountability. There is, however, emerging research in police settings that provides empirically based guidelines for the integration of autonomous systems (Dolata & Schwabe, 2023). Building on this foundation, we focus on the firefighting domain to explore accountability dynamics within socio-technical systems involving autonomous drones.

The lack of field studies examining the use of autonomous drones underscores the critical need for research on their real-world application. Realistic operational scenarios in which participants interact directly with autonomous drones can significantly shift perspectives and provide deeper insights into accountability relationships. Unlike theoretical assumptions or survey-based perceptions, empirical data from field trials uniquely reveal how accountability and operational challenges situatively emerge. Building on this need, Kempton et al. (2023) recommend adopting a comprehensive view of AI accountability. This includes a) the obligation of those involved in AI development and deployment to justify their actions, b) the ability of stakeholders to challenge AI systems, and c) mechanisms for sanctioning unacceptable system outcomes. In response to this call, this paper provides empirical evidence that specifically addresses Kempton et al.'s first recommendation by examining how the fire department's use of autonomous drones affects its obligation to provide accountability. Based on our findings, we offer practical recommendations to improve accountability in the context of autonomous systems.

## Methods

This research formed part of an extensive project employing a user-centered approach to explore the potential applications of autonomous drones in supporting firefighters. To this end, we first conducted discussions with firefighters using predefined questions to generate ideas and suggestions for deploying the autonomous drone system. Following an evaluation of the technical capabilities of drones for a number of proposed applications, we have identified possible scenarios for a field trial. Subsequently, through two field trials involving different volunteer fire departments, we sought to assess firefighters' operational procedures, explore drone usage in identifying fire sources and potential victims, analyze drone interactions, and understand bystanders' and firefighters' perceptions. The objective of the field trials was to gain insights into how firefighters could effectively use autonomous drones during emergency operations, how drone systems should be designed to support firefighters, what benefits and challenges bystanders associate with autonomous drones that interact with them as a part of public emergency service, and how the use of such drones affects the attribution of accountability in these contexts. Two distinct simulation contexts were employed in the exercises: a building and a forest fire. We did not find major differences between the two in terms of participant responses, so we report aggregated results.

## *Field trial*

The field trial included a total of 95 participants divided into two main groups. First, we included a group of firefighters from two volunteer firefighter departments. The firefighters were ranked as general commanders, exercise commanders, or civilian members of the firefighter department. The exercise commander organizes the exercise and ensures the conditions are realistic for the exercise to succeed. The role of the general commander was to lead the entire operation. In addition to the existing task forces, a new role was created. The general commanders received support from the drone officer located next to him. The drone officer accessed the drone's video stream via a digital handheld device, evaluater the drone's information and decisions, and relayed relevant insights to the commander. Operationally, the drone was positioned above the regular firefighters in the command hierarchy, but below the general commander. This setup ensured that while the drone operated semi-autonomously and provided critical real-time data, it remained integrated within the existing command structure through a designated human actor. The exercises were mandatory for all civilian members of the department, though they received limited information about the possible timing of the exercise and the use of drones.

As specified in the scenario, the field trial also included a second group of individuals who were not associated with the firefighting department , who were assigned the role of bystanders, i.e., individuals who happened to be at the fire scene by accident. They were informed that they were participating in a live fire exercise and that novel technologies would be used, but they did not know that drones would be involved. The bystanders were positioned at various locations. To make the situation more realistic and simulate the bystanders' attention level, all participants were given a task unrelated to the fire incident. For example, a small group was asked to play volleyball on the lawn at the backside of the building so they would not initially know what was happening or be primed that drones were being used. Some of the bystanders received special instruction about different levels of injury and mobility (victim role). The main task of these individuals was to remain in an assigned location and wait for rescue. During the field trial, all bystanders were allowed to interact realistically with the environment (including the drones) and cooperate with firefighters. It provided an opportunity to explore the potential and challenges of drone interaction with different user groups. For documentation and evaluation purposes, the course of the field trial was recorded on video from different points of view.

The scenario for the field test was developed and agreed in close cooperation between the authors, a student assistant and two doctoral students, and the firefighters themselves. The test scenario is based on a fire situation either in a building or in a forest. The course of the field trial was as follows: While the bystanders performed their tasks, the exercise commander triggered the field trial by either setting the fire (in the case of a forest fire) or starting the smoke machine (in the case of a building fire). A few minutes later, information about the fire reached the fire station, and the general commander triggered the alarm, whereupon the entire crew was mobilized. The command center initiated the operation by quickly ordering the drones. The drones left the base at approximately 1500 meters away and provided the first images of the emergency location and information about the accident. Just one minute after the alert, the drones were airborne. It took the civilian members of the fire department another 15-20 minutes to arrive on the scene. Two drones were used in this operation, acting as a team. While one drone provided an overview from above and communicated with bystanders and victims via loudspeakers and spotlights, the second drone conducted a detailed search for victims inside the building or in the forest. The operation was terminated after all victims were rescued, the fire was extinguished, and all possible residual members were removed. Figure 2 shows the modified organization of the fire scene.

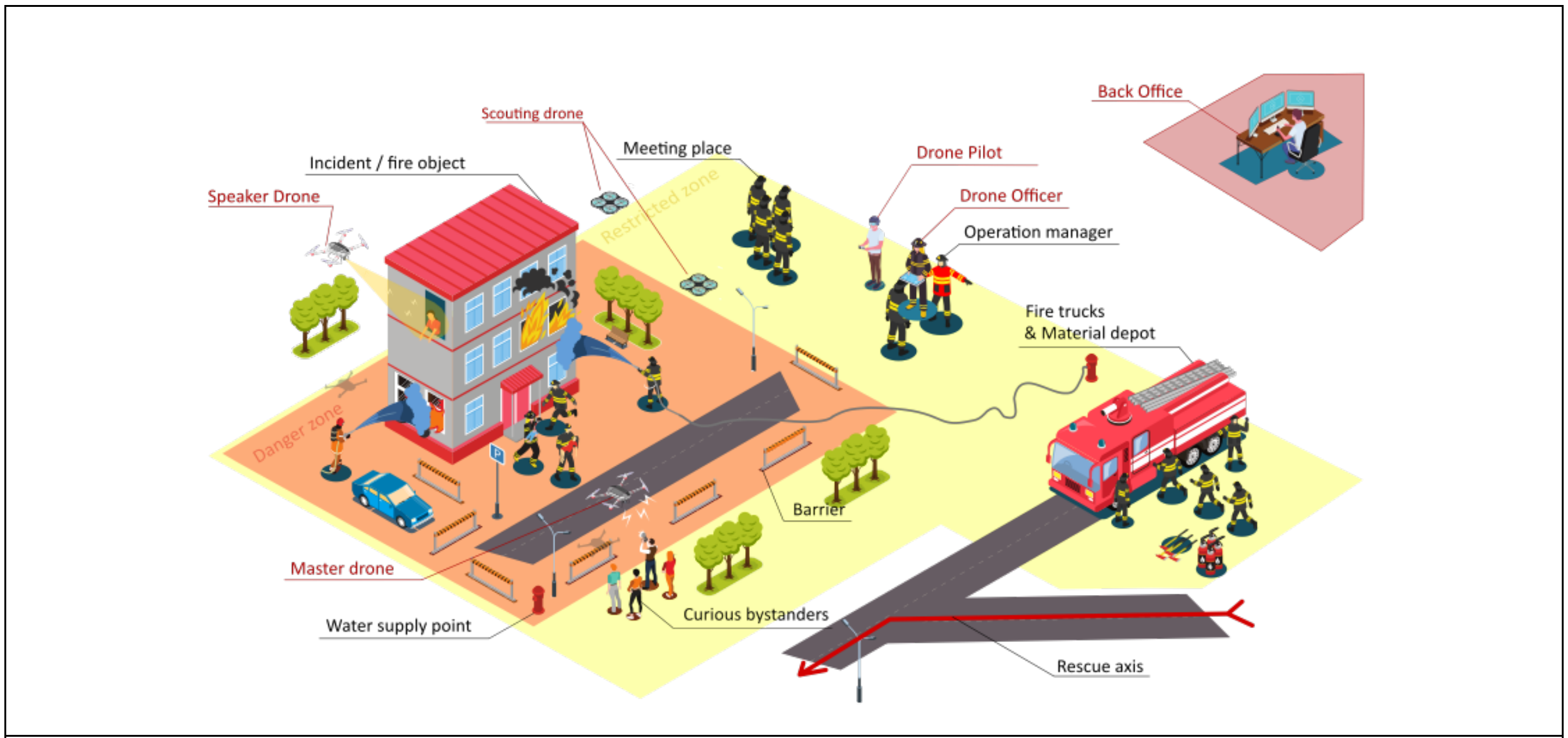

**Figure 2. Firefighting operation involving multiple (semi-) autonomous drones (Li et al., 2024)**

The following preselected scenarios for human-drone interaction were integrated seamlessly into the firefighters' regular exercises to replicate real-life conditions as closely as possible:

(1) Drones verify the correct address and confirm the fire location
(2) Drones obtain information about the surroundings, incl. determining the extent of the fire
(3) Drones contact third parties, send them away from access roads, accesses, rescue lanes and water resources
(4) Continuous information gathering and evaluation of future events
(5) Support and search for persons (incl. victims) within the emergency area
(6) End of operation and documentation of incidents

The selected scenarios were based on an analysis of previous operations by the fire brigade, which showed that the drone could reach the fire's source faster than the first fire unit. As a result, the drone was the first and only representative of the firefighting team on the scene in the initial phase.

To cultivate the illusion of complete autonomy of the drone during our study, we used a Wizard-of-Oz (WoZ) technique following previous methods in drone interaction research (Cauchard et al., 2015; Karjalainen et al., 2017; Dahlbäck et al., 1993). This approach allowed us to carefully monitor and regulate the drone's actions while ensuring the safety of all participants. By concealing the presence of the human operator behind a curtain, we created the impression among participants of an independently functioning drone. This intentional concealment allowed for an unbiased examination of participants' interactions with the technology.

### *Participants and Data Collection*

A total of 67 firefighters participated in both exercises. The firefighters' participation in the exercises was mandatory and not remunerated. Since drones are becoming increasingly widespread, it is reasonable to assume that some firefighters have come into contact with them in both private and professional contexts. Some of the firefighters had prior experience using drones in emergency situations; they operated and used drones to investigate emergencies in the field. The research team recruited 28 volunteers to act as bystanders using the snowball method (word-of-mouth) and through direct contact with the participating fire departments. Bystanders were required to speak either English or German at a communicative level, with demographical distribution roughly representing the population living in Switzerland. For approximately 2 hours of participation, bystanders received compensation equivalent to 80 USD, in line with the university practice.

Directly after the field trial (within the first one to two hours), participants underwent a brief individual interview with a member of the research team on site. The interview was semi-structured and covered five main topics: the overall impression, drone performance perception, the interaction with the drones, drone autonomy, accountability, and the responsibility related to the actions of a drone. Furthermore, all subjects received one particular question: Who should be accountable in case of failure? We focused on the topics for two main reasons. The topics were identified upfront as particularly relevant in prior literature (Dolata & Schwabe, 2023). Additionally, the focused character of the interviews allowed for efficient collection of data, such that participants did not have to wait. Despite our best efforts, it was unfeasible to interview all participants, but the analysis of data indicates that we achieved saturation. Specifically, statements related to the selected topics appeared repetitively without leaving major aspects unexplored. For the purposes of this study, we focus mainly on the questions of accountability related to the actions of a drone. Since it was a volunteer fire department, the exercises were conducted in the evening to allow as many firefighters as possible to participate. However, this had a limitation for the planned interviews, as not all subjects had time for an interview after the exercise. Nevertheless, we conducted a total of 50 interviews in both field trials. Following Field Trial 1, the interviews with bystanders lasted an average of 12.7 minutes, while those with firefighters averaged 18 minutes. In the subsequent Field Trial 2, the interview durations increased to an average of 17.45 minutes for bystanders and 19 minutes for firefighters.

### *Data analysis*

The recorded interviews were transcribed following the intelligent verbatim standard and translated into English. The transcripts were analyzed according to the exploratory-interpretive paradigm (Saldana, 2021; Stebbins, 2001), meaning that the goal was to reach a shared interpretation of the data rather than an objectively "correct" one (Merriam & Grenier, 2019). The analysis of the data was done in two steps. First, a single researcher identified the dominant themes in a bottom-up process. Secondly, the authors undertook an iterative sense-making approach in which an additional level of coding was added specifically focusing on accountability. This step consisted of two interpretation workshops involving all the authors, which allowed them to assess the completeness of their interpretation and enrich the range of possible meanings. While objectivity in such a task is difficult to achieve and not always desired, this process ensured that the results did not emerge solely from an authors' individual, subjective perspectives but from an inter-subjective, interpretative process. However, potential interpretive limitations and individual biases may remain and are acknowledged.

Once the data analysis was completed, the research team held an internal workshop to jointly identify accountability challenges and discuss potential mitigation solutions, which were then developed into concrete policy recommendations. The discussion was guided by the results of the interview analysis, with particular emphasis on the identified codes. In addition, we drew inspiration from related guidelines developed in the context of policing (Dolata & Schwabe, 2023) and used them as a reference point to further develop and contextualize our recommendations for the firefighting use case. This approach ensured that our policy guidelines are both empirically grounded and informed by existing best practices.

## Results

### *Hierarchy and Human Accountability*

Our study is of an exploratory nature and focuses on the impact of the use of autonomous drones on accountability from two different perspectives: firefighters and bystanders. In the interviews, we explicitly asked study participants who should be held accountable in an event of failure, e.g., when the decision taken by the drone would create unexpected danger. During the analysis, it became clear that although all participants were asked this question, some provided evasive or contradictory responses. We decided to remove them (n=4) from this analysis (though we retained them for the qualitative analysis reported below). This explains the lower number compared to the total number of participants. We decided to first provide a quantitative evaluation of this question to illustrate the overall tendency. These frequencies reflect the number of coded responses within our sample and should not be interpreted as statistically representative. Figure 3 summarizes the frequency of answers to the question of who is accountable if a firefighting operation results in a loss that can be linked to the drone's actions.

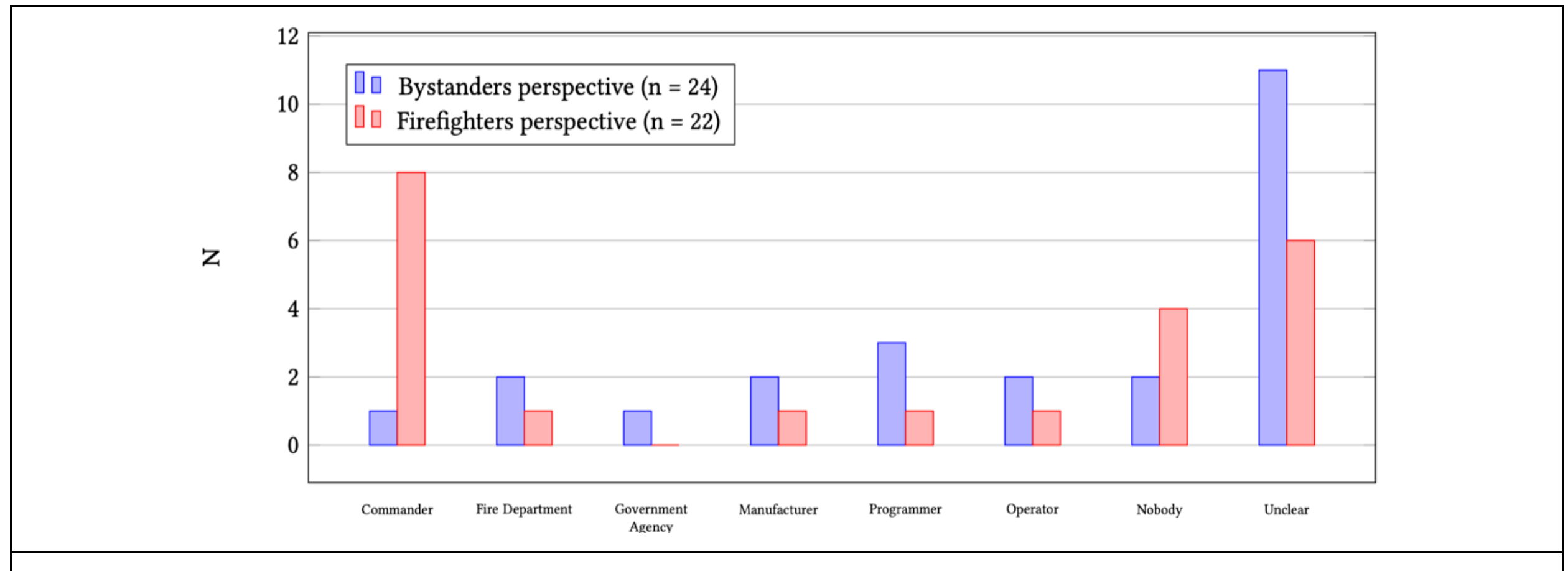

**Figure 3. Distribution of respondents' answers to the question "Who is accountable in case of failure?"**

In particular, most firefighters tend to attribute accountability to the commander following the chain of command in the fire departments. This implies that they employ a notion of strict liability, assigning accountability to the person highest in the hierarchy, who, according to the interviewees, is considered vicariously accountable for outcomes related to the drone. In contrast, many of those who declare nobody to be accountable employ the "good-will frees blame" principle, which implies that they think about accountability in terms of blameworthiness. In other words, sometimes, accidents may happen for which no one can be blamed and hence held accountable. Further, many are uncertain about the appropriate attribution of accountability. The bystanders are mostly uncertain as to who might be accountable for the failure of a drone application. Many also see the producer (including the manufacturer and developer of the drone) as the responsible entity.

### *Identification of Drone as Part of Organization*

Other accountability attributions seem to be equally distributed across the other group. It was striking, however, that a significant number of participants did not connect the drone to the firefighting department because of its appearance. Therefore, the drone wasn't perceived as an official actor: *I just miss the blue light in this exercise [...] without that, the drone has no authority [...] if the drone just comes in black, then it's ... unsure if it belongs to an official organization.* (B15, B16, B17, B18; group discussion[1]).

At least the majority of participants attributed responsibility to the collective of the fire department or other organizational units rather than individuals: "*the organization that used the tool will have to take responsibility. But I don't think it will be the drone pilot himself unless he can be proven to have been very negligent*" (F-34). This might be reportedly because *"Ultimately, the supply strategy lies in the hands of human structure"* (sic) (F-33). The firefighters quoted in these last passages tend to shift accountability away from the drones. Sometimes, the respondents handled this dispersion of accountability with some degree of conceptual confusion. Compare, for instance, *"and if they say, 'Yes we want to use the drone,' and then something happens. Yeah, then it's kind of the fire department's fault"* (B-13; emphasis added), as opposed to *„the firefighters [...] should have taken over"* (B-13; emphasis added). Such inconsistencies reflect the complexity and uncertainty surrounding accountability when drones are deployed in firefighting operations.

### *Perceived Autonomy vs Accountability*

To explore a potential link between the perceived autonomy of the drone and the attribution of accountability to the drone, we also collected insights on autonomy ascription by asking participants to rate the drone's perceived ability to make decisions independently of a human on a scale from one to five (with five indicating the highest value). Yet frequently, many participants expressed normative views. That is, they did not report their perception of the drones in the particular scenario but rather communicated their opinion about what should be the case, e.g., "*I think ultimately a human being should still make the decisions*" (F-12). Others evaded the question with statements such as *"I think it certainly will come."* When questioned again, several participants struggled to assign a number. Many firefighters knew which parts about the drones were autonomous and which tasks were mocked. We excluded such answers (n=22) from the quantitative analysis but kept them for qualitative exploration. Even if not complete, the data provides an indication of the general tendency. Figure 4 illustrates the results concerning the perceived ability of drones to make their own decisions based on the respective interviews undertaken.

[1] The first letter of participants' ID's indicate their role. F stands for firefighter, B stands for bystander.

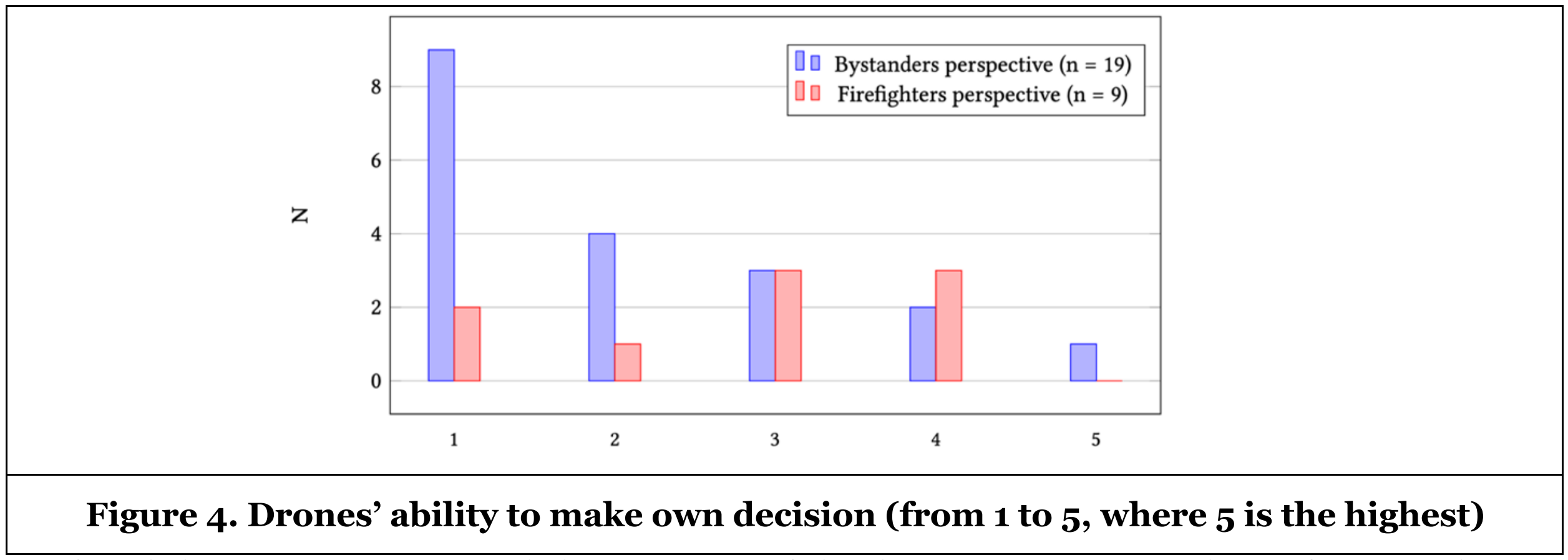

**Figure 4. Drones' ability to make own decision (from 1 to 5, where 5 is the highest)**

We found that participants struggled to conceive of the drones as genuine authors of actions that are completely independent from human input. On some occasions, the drone is simply equated to *„another arm that can be far from you, and you can control it"* (B-14). More elaborately: *„if it had flown autonomously now, it wouldn't be doing something independently only what it's told to do. So, from that point of view actually the drone is not at all acting independently"* (B-13). Such considerations in our data set suggest a broader reluctance among respondents to consider autonomous drones as being on a par with human firefighters in terms of accountability attribution.

Interestingly, participants tended to ascribe accountability to human agents – either collective or individuals – rather than to the drones themselves, even when our interviewees actually hypothesized autonomous drone capabilities: *„I mean, even if the drone flies autonomously, someone has to send the drone into the air [...] And so there must be someone who, even if it flies autonomously, I still assume that there must be a drone pilot who can intervene in an emergency."* (F-33). Again, *„[the operator] actually has to know what the drone can do and what it can't do"* (B-13). Thirdly, *„So, it's of course easy to say the drone is guilty, right? It's a technical error. [...] Yes, but then the manufacturer [would be responsible]"* (B-23). These quotes, drawn from a firefighter and a bystander in our field trial scenario, illustrate different rationales for assigning blame to different actors. Yet, they share a common theme: They all seem to shift away accountability from the drone itself in cases of malfunctioning or wrong decisions. Typically, the operators – responsible for the manual operation of the drones – are viewed as ultimately accountable, even when it is the machines' autonomous operations that result in undesired outcomes. This reflects a broader consensus that humans must stay in the loop to remain accountable in extreme cases: "*Ultimately, it's always a human failure. So, it's now kind of, the drone is giving some suggestion, but you still have the opportunity to decide, yes or no. [..]. But you [human operator] as the clever one should actually be able to say yes or no"* (F-9). Otherwise, accountability is still transferred to human components, as done in the last quoted passage, and in particular to the manufacturer of the drone in the occurrence of 'technical errors'.

## Discussion

With our goal of identifying specific accountability challenges in the context of autonomous drone-supported firefighting and exploring strategies to mitigate them, our findings paint a complex picture. There is generally a tendency to attribute accountability to the fire department as a collective entity or to individual humans, such as the commander, operator, or manufacturer. When respondents attributed independent decision-making capacity to the drone, they tended to shift accountability away from particular individuals involved in the operation (e.g., drone operators or other firefighters) and were more inclined to place accountability vicariously on collective entities (e.g., the fire department as an institution, the incident commander as its representative, or the manufacturing company). Interestingly, none of the interviewees identified the drone itself as an accountable entity. Initially, we hypothesized that accountability attribution would likely correlate with people's perception of the drone's autonomy. However, even when participants contemplated autonomous drones, accountability for errors or malfunctions was still attributed to other (human) actors.

From the perspective of accountability theory, it is essential to distinguish between attributing accountability to certain individual agents, such as firefighters, and the institution they represent, i.e., the fire department (Bovens, 2007; Mulgan, 2003). The collective accountability of the fire department may extend beyond the firefighters directly involved in the rescue operation, potentially reaching superiors or higher authorities. In this case, the question is not so much about whose action led to the drone malfunctioning, but rather who authorized its deployment and in whose name the drone is being used. In this case, accountability might be rather attributed on the basis of institutional roles and authority rather than direct causal involvement (cf. Mulgan, 2003, chpt. 6). By contrast, when accountability is attributed to a group such as 'the firefighters', the focus tends to rest on the actions of individuals at the scene, regardless of whether their actions were coordinated. This creates difficulties when an autonomous drone fails in ways that cannot be causally traced to identifiable misconduct by particular firefighters.

Generally, our interviews did not echo the accountability-gap concerns as discussed in the academic literature (Binns, 2018; Kroll, 2020; Matthias, 2004; Nissenbaum, 1996). One possible explanation is a lack of awareness for this issue among our participants. In view of the increasing complexity of newer generation AI models coupled with their lack of explainability, it is highly debatable whether the actors to whom our interviewees assigned accountability to – such as commanders, operators, or manufacturers – can actually ensure the drone's controllability, provide explanations in the event of failure, and conduct thorough investigations to prevent similar events in the future. One could conclude from our interviews that the clear tendency to hold humans accountable for drones' actions coupled with the repeated call for human control over drones suggests an implicit intuition that drones should only take over tasks that humans can in fact control and account for – an idea captured in the academic literature under the notion of "meaningful human control" (Santoni de Sio & van den Hoven, 2018). In line with broader discussions in the AI ethics literature, this points not only to the importance of transparent and scrutable algorithmic models (Binns, 2018; Cobbe et al., 2021), but also the need for adequate technical literacy among firefighters, including a thorough understanding of the capabilities and limitations of drone technology.

We associate the great uncertainty concerning accountability attribution among bystanders to the fact that they are less aware about the fire brigade's internal decision structures and responsibility distribution. Nonetheless, even when asked about their intuitions, the uncertainty remained. This could indicate that autonomous drones, as new technologies, bring to the surface new ethical challenges for which lay persons have not yet formed clear intuitions. Conversely, for firefighters it seemed more obvious to attribute accountability to commanders, probably reflecting the existing hierarchical organizational culture in fire departments. For these firefighters, it seemed no big challenge to integrate drones into existing accountability frameworks. Nonetheless, the fact that the second and third largest group of fire fighters answered either 'unclear' or 'nobody', gives reason to conclude that ambiguity persists as to how accountability concerning drone use is distributed. Such ambiguity must be resolved if drones are to be responsibly deployed. From the perspective of both, responsible AI governance and democratic legitimacy of public institutions, it is essential to specify who must account for what, and under which circumstances, in order to avoid accountability-gaps (Thompson, 2017).

Another interesting finding is the fact that, when asked during the interviews, bystanders considered firefighting brigades vicariously accountable for drone actions, but at the same time, many bystander participants did not recognize the drones as official representatives of the fire brigade during the field trials. This leads to uncertainty about what to expect from drones in terms of accountability when they are operated on site. Just as firefighters' uniforms signify their role as representatives of an official, accountable institution, drones should similarly carry identifiable labels to be clearly linked to their operating institution.

The reluctance to attribute independent decision-making capabilities to autonomous drones appears to be shaped by factors such as the inherent nature of drones and the development of autonomous technology. Many autonomous drones are still operated under human supervision and control, contributing to the perception that autonomy is not absolute. Public perception plays a crucial role in the widespread acceptance of autonomous technologies. Skepticism or lack of trust in the capabilities of autonomous drones can hinder the willingness to delegate independent decision-making powers and responsibilities. Yet, participants' general reluctance may not be fixed. The inclination to view drones as fully autonomous agents who can be held accountable may increase over time as technology matures and public confidence in its capabilities grows.

Our findings make several important contributions to the IS literature. While previous ERIS research has primarily focused on the design and adoption of traditional information systems in emergency contexts (Beydoun et al., 2018; Prasanna et al., 2017; Turoff et al., 2004) our study provides empirical evidence of the complex sociotechnical challenges that emerge when integrating truly autonomous agents into emergency operations. The field trials reveal that drones, as examples of on-site dynamic information collection systems (Yang et al., 2009), introduce unique coordination and communication challenges that extend beyond traditional human-computer interaction paradigms. Specifically, our findings demonstrate that the integration of autonomous systems into established emergency response workflows requires careful attention to organizational identity, hierarchical structures, and role clarity – factors that are often overlooked in prior ERIS research.The uncertainty observed among both firefighters and bystanders regarding drone authority and capabilities indicates that is possible that addressing those aspects might contribute not only to external acceptance of ERIS but also to their adoption by the members of emergency services. IS should not only consider technical advantages of dynamic and autonomous systems in emergencies, but also consider their social and organizational embedding. This empirical evidence extends current IS understanding of emergency response systems by showing how autonomous agents can simultaneously enhance operational capabilities while creating new forms of accountability complexity that must be addressed for successful adoption. This adds to the research on public's perceptions of drones in emergency situations (Dolata & Aleya, 2022; Dolata & Schwabe, 2023) and advantages of dynamic and autonomous extensions to existing ERIS (Yang et al., 2009).

Additionally, this study advances IS research on accountability and AI accountability by providing rare empirical evidence of how accountability is actually attributed in high-stakes human-AI collaborative scenarios. While previous IS accountability research has largely focused on policy violations and compliance in organizational settings (Dubnick & O'Kelly, 2019, 2019; Sjöström et al., 2022; Vance et al., 2013) our findings reveal the complex dynamics of accountability attribution when autonomous systems operate as organizational representatives in service contexts. The observation that participants consistently refused to attribute accountability to drones themselves, instead shifting responsibility to human actors or organizational entities, provides crucial empirical support for theoretical concerns and speculations about the complex nature of accountability stated in more recent research (Kempton et al., 2023; Nguyen et al., 2024). More importantly, our findings demonstrate that accountability challenges in human-AI teams emerge not primarily from technical opacity or algorithmic complexity – the focus of much IS AI accountability research (Kempton et al., 2023; Kim et al., 2020) – but from organizational ambiguity, role confusion, and inadequate communication of system capabilities. This also confirms prior conceptual research speculating that humans do not perceive AI as an individual, detached entity but rather as embedded in an activity system which embraces the social and material context as well (Dolata et al., 2023).

### *Autonomous Drones Accountability Policy Guidelines*

Addressing accountability challenges (RQ2) that arise within complex socio-technical systems is essential to ensure the effective and legitimate integration of autonomous drones into firefighting operations. Our findings highlight several key areas that complicate the attribution of accountability. To mitigate these challenges, we propose a set of empirically grounded policy guidelines that aim to clarify accountability and, as a result, promote trust in the use of autonomous drones in the long run. These guidelines serve as practical recommendations for policy makers and firefighting organizations to ensure that accountability is maintained at all levels of deployment.

> ***Organizational Affiliation of Autonomous Drones**: Firefighting organizations must ensure that drones are visibly and clearly recognized as part of the organization. This not only*

> *flags them as part of an official institution promoting trust, but also reinforces their role as accountable actors in emergency operations*

Many participants in our study did not associate drones with the fire service due to their appearance, which led to confusion about accountability. To reduce such ambiguity, drones should be clearly branded or marked to signal their affiliation with the fire department. Ensuring that drones are visibly recognized as part of the firefighter team facilitates their integration into existing accountability frameworks. This visual and communicative association is crucial for both bystanders and operators to perceive the drone as a legitimate actor within the system, following Bovens' (2007) accountability framework.

> ***Communication of Autonomous Capabilities****: The autonomous and intelligent capabilities of drones must be clearly communicated to all parties.*

Although drones may operate autonomously, many participants in our field trials attributed accountability solely to human operators, often unaware of the drone's autonomous functions. Clearly communicating the autonomous and intelligent capabilities of drones is essential to ensuring accurate accountability attribution. By highlighting the drone's ability to act independently on behalf of the fire service, fire services can adjust the expectations of victims, bystanders, and other stakeholders. Transparency in the drone's decision-making processes can foster trust, facilitate cooperation, and enable operations to begin earlier, even before human responders arrive on site. This not only prioritizes the accountability of human actors and the organization but also ensures that accountability is maintained even in the absence of immediate human presence. By embedding drones as visible and trusted components of firefighting operations, their role as autonomous actors can be better understood and accepted.

> ***Clear division of labor****: A well-defined allocation of tasks and responsibilities between drones and human operators must be established and communicated to all stakeholders. This clarity ensures that accountability is appropriately assigned during operations, reducing confusion and enhancing operational efficiency*

In the field trials, participants frequently held human operators accountable, even when drones were capable of performing tasks independently. This highlights the critical importance of ensuring that operators maintain control over and knowledge of drone actions. Firefighting organizations must establish and communicate a clear division of tasks and responsibilities between drones and human operators. This clarity enables operators to monitor drone activities effectively, explain the drone's actions, and justify its use for specific tasks. By ensuring that the use of drones is reasonable and well-assessed, accountability can be accurately attributed to the supervising human operator or team, rather than becoming diffuse or misplaced. This approach safeguards compliance with established standards, ensures transparency, and reinforces trust in the integration of autonomous systems within emergency response workflows.

> ***Explainability of autonomous drones:*** *Human operators must have sufficient understanding of the capabilities, limitations, and decision-making processes of autonomous drones to provide clear and reasonable explanations for their actions*

Operators cannot delegate tasks to drones without maintaining an understanding of the drone's behavior. In line with Bovens' model of accountability (2007), operators must be able to explain the drone's actions, whether it is during operations or in a formal review setting. This not only ensures that operators make informed decisions, but also that they can justify those decisions to accountability forums. The ability to explain a drone's actions builds trust and clarity within the system and ensures that human accountability is aligned with autonomous operations.

> ***Transparency between manufacturers and operators:*** *A transparent and close relationship between drone manufacturers and fire services is essential to maintain accountability*

Transparency and stable communication channel between manufacturers and deployer is crucial for identifying and explaining potential malfunctions or limitations of autonomous drones. This relationship ensures that, in the event of a failure, both the fire service and the manufacturer can review the issue and implement corrective measures. It also enables fire services to report any system limitations or incidents to oversight bodies or the public, supporting the accountability process. This type of collaboration is key for ensuring long-term safety and preventing future incidents through shared knowledge.

> ***Internal handling of accountability**: Fire services must establish internal processes to investigate failures and enforce accountability when necessary*

Robust internal procedures are critical for taking accountability seriously within firefighting organizations. These processes must include a systematic approach to investigating incidents, including the identification of errors sources and the enforcement of sanctions when standards have been breached. By creating these structures, fire departments can address accountability challenges effectively and ensure that responsible parties are held to account. Fostering an accountability culture strengthens the integrity of emergency operations and promotes public trust in the organization.

> ***Clarification of accountability for group agents**: When assigning accountability to institutions or group agents, procedures must ensure that individuals within the group remain accountable*

Assigning accountability to abstract entities, such as institutions or autonomous systems, risks obscuring who is ultimately accountable for particular actions, whether individual or collective. This mirrors the broader challenge of holding group agents accountable. To avoid such diffusion of responsibility, fire services should establish procedures that clearly assign roles within the organization and ensure that individuals remain accountable for the outcomes linked to those roles. These procedures should include mechanisms for internal investigations, ensuring that responsibility is traceable to individuals, thereby maintaining clear accountability even within complex organizational structures.

There guidelines developed here aim to provide policymakers with a structured approach to addressing accountability challenges in the use of autonomous drones within firefighting operations. They highlight the importance of clarity in roles, explicit institutional procedures, and transparency to ensure that accountability is maintained at all levels.

# Conclusion

This paper investigated accountability challenges in sociotechnical settings where firefighting operations are supported by autonomous drones and formulated concrete policy guidelines to mitigate these issues. We observed considerable uncertainty regarding the attribution of accountability in relation to the organizational use of drones. Notably, individuals do not attribute accountability to drones themselves. Instead, accountability tended to be shifted to collective actors, such as the fire department or the firefighters as a group. As highlighted in the introduction paper, the use of autonomous drones is expanding across various public and private sectors, including agriculture and law enforcement. Our findings are therefore pertinent for assessing how applications of drone technology, and the perceptions surrounding it, may influence public acceptance and perceptions of institutional legitimacy.

# Limitations and Future Work

This study has several limitations that should be acknowledged. First, the empirical data was collected in the context of simulated emergency exercises rather than real crisis events. While the scenarios were developed and conducted in close collaboration with fire departments to ensure realism, the fact that they were framed as exercises may have diminished participants' perceived sense of urgency and risk. This lower level of situational pressure could have influenced their cognitive and emotional responses, including how they attributed accountability in moments of ambiguity or system failure. As such, the external validity of the findings is necessarily constrained. Second, participants' responses were gathered through post-hoc interviews conducted after the field trial. While this approach allowed for in-depth reflection, it also carries the risk of retrospective rationalization. Immediate, in-situ intuitions and reactions, particularly those related to accountability, responsibility, and perceived agency, may not have been fully captured. Future studies could benefit from real-time methods to better access participants' intuitive reactions during interaction with autonomous systems.

These limitations reflect the unavoidable trade-offs in designing empirical field studies involving autonomous systems in sensitive, high-risk domains. Safety, control, and realism must be balanced against the constraints of operating in real-world environments. Nonetheless, we believe that our methodological choices provide a credible and meaningful foundation for further research.

Looking ahead, future studies could explore drone configurations that embody greater degrees of operational autonomy and evaluate how such designs affect perceptions of accountability and agency. Comparative designs that vary command structures, visual cues, or communication modalities may yield deeper insights into human-drone collaboration. This study offers an empirical perspective through which theoretical assumptions about responsibility in human-agent systems can be examined. We see potential for interdisciplinary dialogue between philosophy, human-computer interaction, and information systems to refine conceptual frameworks and inform public policy around the deployment of autonomous technologies in critical public services.

## Acknowledgment

This research was a part of the project titled 'Automated Decision-Making and the Foundations of Political Authority,' financially supported by the Swiss National Science Foundation (project number 208013). We would like to convey our sincere appreciation to all project members for their constructive feedback and active participation throughout the development and evaluation phases of the drone system. Special acknowledgment is extended to the participants in the field trials for generously sharing their insights and feedback. Our gratitude goes to the Altdorf and Silenen fire stations and the dedicated firefighters involved in the field trials for their steadfast support and collaboration during the exercises. We would also like to thank the University of Zurich members – Andreas Bucher, Sven Eckhardt, Yiwei Wu, Beat Furrer, Simon Giesch, Yiming Hu, Yinglun Liu, Moyi Li and Pascal Marty – for their indispensable assistance in data collection and analysis. Furthermore, we express our thanks to the anonymous participants in our study and the review team for their invaluable insights and guidance in refining this manuscript.